\documentclass[lettersize,journal]{IEEEtran}

\usepackage{amsmath,amsfonts}
\usepackage{array}
\usepackage[caption=false,font=normalsize,labelfont=sf,textfont=sf]{subfig}
\usepackage{textcomp}
\usepackage{stfloats}
\usepackage{url}
\usepackage{verbatim}
\usepackage{graphicx}
\usepackage{cite}

\usepackage{booktabs}
\usepackage{xcolor}
\usepackage{colortbl}
\usepackage{multirow}
\usepackage{algorithm}
\usepackage{algorithmic}
\usepackage{tikz}
\usetikzlibrary{calc,shadows}
\usepackage{ragged2e}
\usepackage{enumitem}
\usepackage{placeins}
\usepackage{xurl}
\usepackage{hyperref}

\begin{document}
\bstctlcite{IEEEexample:BSTcontrol}

\title{VisualRouter: Query-Grounded Visual Sampling for\\ 
Long Video Understanding}

\author{
Haiyue~Zhang,
Yi~Bin,
Xun~Jiang,
Zeyu Ma,
Duo~Peng,
Guoqing~Wang,
Yang~Yang,~\IEEEmembership{Senior~Member,~IEEE,}
and~Heng~Tao~Shen,~\IEEEmembership{Fellow,~IEEE}%

\thanks{Haiyue Zhang, Yi Bin, Xun Jiang,  Duo Peng, and Heng Tao Shen
are with Tongji University, Shanghai, China.
(Corresponding author: Yi Bin, e-mail: yi.bin@hotmail.com).}

\thanks{Zeyu Ma, Guoqing Wang, and Yang Yang are with the University of
Electronic Science and Technology of China, Chengdu, China.}
}

\markboth{Journal of \LaTeX\ Class Files,~Vol.~14, No.~8, August~2021}%
{Shell \MakeLowercase{\textit{et al.}}: A Sample Article Using IEEEtran.cls for IEEE Journals}

\IEEEpubid{0000--0000/00\$00.00~\copyright~2021 IEEE}

\maketitle
\begin{abstract}
Large vision-language models (LVLMs) have achieved significant progress in video understanding, yet understanding long videos remains challenging due to the large number of visual tokens and limited context windows. Visual sampling provides a practical solution by selecting an informative subset of frames. However, existing methods typically either rely on relevance-aware sampling, leading to redundant frame selection and insufficient temporal coverage, or adopt a fixed sampling strategy regardless of query type. In this paper, we propose VisualRouter, a training-free and plug-and-play framework for query-grounded visual sampling. VisualRouter first classifies each query as either global or local and then applies the corresponding sampling strategy. For global queries, it employs a relevance-coverage hybrid strategy that preserves temporal coverage while retaining query-relevant visual evidence. For local queries, it adopts an event-aware frame selection strategy that performs event partitioning, segment-level frame allocation, and intra-event frame selection, jointly balancing relevance, coverage, and diversity with a limited number of input frames. Experiments show that VisualRouter consistently improves multiple LVLMs over uniform sampling, achieving gains of 5.2\%, 7.7\%, and 11.6\% on Video-MME, LongVideoBench, and MLVU with Qwen2.5-VL-7B, and outperforming existing training-free visual sampling methods under the same setting. Code is available in this \href{https://github.com/ZHYJYJYJY/VisualRouter}{URL}.
\end{abstract}

\begin{IEEEkeywords}
Long video understanding, large vision-language models, query-grounded visual sampling.
\end{IEEEkeywords}

\section{Introduction}
\IEEEPARstart{L}{arge} Vision-Language Models (LVLMs) have achieved remarkable progress in image and short video understanding, and have recently been extended to long video understanding~\cite{maaz2024videochatgpt,zhang2025llavavideo,li2024llavaonevision,bai2025qwen25vl,lin2023videollava}. Compared with short videos, long videos cover broader temporal spans, involve more complex event structures, and contain substantial visual redundancy~\cite{fu2025videomme,wu2024longvideobench,zhou2025mlvu}. Long video understanding requires LVLMs to reason over extended temporal contexts, capture event progression and cross-segment dependencies, and retrieve query-relevant visual evidence from highly redundant visual inputs. However, directly feeding long videos into LVLMs remains challenging because the resulting massive number of visual tokens quickly exceeds the limited context windows of existing models.

\begin{figure}[!t]
\centering
\includegraphics[width=2.6in]{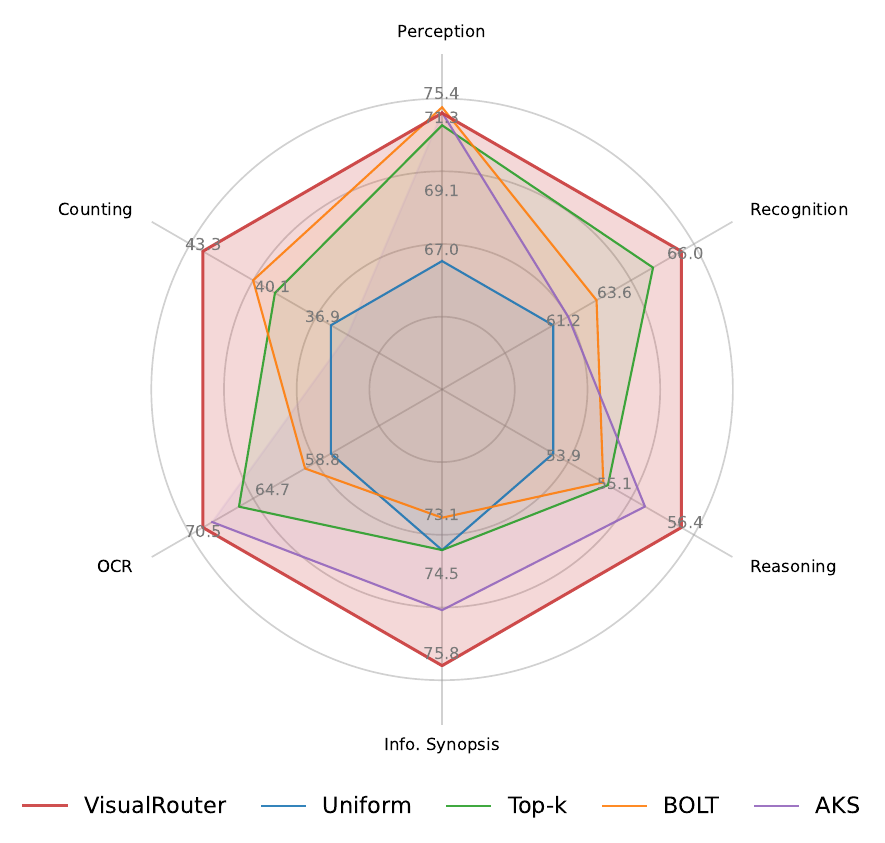}
\caption{Accuracy(\%) of visual sampling methods across six question categories in Video-MME, evaluated with Qwen2.5-VL-7B and 16 frames.}
\label{fig:question_type}
\end{figure}

To address this challenge, existing studies mainly follow two directions. The first focuses on the model or token level, reducing visual tokens by token compression, visual memory, or long-context extension~\cite{xu2025slowfastllava15,zhang2024longva}. Although effective, these methods typically require additional training or architectural modifications, which limits their applicability as general preprocessing strategies across different LVLMs. The second performs visual sampling at the input level, retaining only frames that are informative with respect to the query and video content~\cite{tang2025aks,zhang2025qframe,liu2025bolt,sun2025mdp3}. This approach offers lower computational overhead and greater deployment flexibility, as it can be applied without retraining the LVLM or altering the underlying model architecture. A widely used strategy is uniform sampling, which maintains temporal coverage but ignores the query-relevant evidence requirements. Consequently, uniform sampling may spend much of the frame budget on irrelevant or redundant video segments, while missing critical visual evidence required for reasoning. Recent studies have used query-frame relevance to guide visual sampling. Some methods select the Top-$K$ frames with the highest relevance scores~\cite{zhang2025qframe,liang2024keyvideollm}, while others combine high-relevance selection with temporal coverage to improve evidence coverage~\cite{tang2025aks}. However, broad temporal coverage does not necessarily reduce visual redundancy, because frames from different timestamps may still contain similar content. To mitigate this issue, some studies incorporate diversity-aware objectives into the Visual Sampling~\cite{li2025maxinfo,zhang2025adardkey}.
\IEEEpubidadjcol

Despite these advances, existing visual sampling methods still face two key limitations. First, query-frame relevance can become unreliable when the query lacks explicit visual anchors, such as in sequence, plot, genre, or video-level summary questions, where the query does not map to specific objects, actions, or individual frames. Second, these methods often optimize query-frame relevance, temporal coverage, or diversity separately, resulting in selected frames that may be relevant but redundant, temporally covered but uninformative, or diverse but query-irrelevant. 

These limitations suggest that visual sampling should adapt to the evidence required by each query. In this paper, we classify queries into two types: global and local. Global queries concern abstract, narrative, or video-level information that cannot be localized to a specific frame or event segment. For such queries, the sampling strategy should preserve broad temporal coverage while retaining query-relevant visual evidence. In contrast, local queries can be grounded in specific objects, actions, scenes, or events, with the supporting evidence typically concentrated within one or a few event segments. For such queries, the sampling strategy should focus on relevant segments, retain sufficient evidence, and reduce visual redundancy. These distinct evidence requirements motivate the query gating mechanism in VisualRouter, which activates the sampling branch best suited to each query.

Motivated by this observation, we propose VisualRouter, a training-free and plug-and-play framework for query-grounded visual sampling in long video understanding. The framework first classifies each query as global or local and then applies the corresponding sampling strategy. For global queries, it adopts a relevance-coverage hybrid strategy that combines relevance-aware Top-$K$ selection with coverage-aware uniform sampling, preserving broad temporal coverage while retaining query-relevant visual evidence. For local queries, VisualRouter follows an event-aware pipeline that partitions the video into event segments, allocates frames according to segment-level relevance, and selects query-relevant and non-redundant frames within each segment. 

The main contributions are summarized as follows:
\begin{itemize}
\item We propose VisualRouter, a query-grounded visual sampling framework that routes queries as global or local based on whether their required evidence is distributed across the video or localized to specific frames, and applies the corresponding sampling strategy.

\item We design two complementary sampling branches: a relevance-coverage hybrid strategy for global queries and an event-aware strategy for local queries that adaptively balances relevance, coverage, and diversity.

\item Experiments on Video-MME, LongVideoBench, and MLVU show that VisualRouter consistently improves the performance of multiple LVLMs over uniform sampling and outperforms existing training-free visual sampling methods under the same evaluation settings.
\end{itemize}

\section{Related Work}
\subsection{Video Large Vision-Language Models}
LVLMs have been extended from image understanding to video understanding, showing strong performance in video question answering, video captioning, and video dialogue. Early architectures, such as Video-ChatGPT~\cite{maaz2024videochatgpt} and LLaVA-Video~\cite{zhang2025llavavideo}, typically sample sparse video frames and project visual features into the LLM space to process dynamic visual content. More recent studies, including LLaVA-OneVision~\cite{li2024llavaonevision}, Qwen2.5-VL~\cite{bai2025qwen25vl}, InternVideo2.5~\cite{wang2025internvideo25}, and VideoLLaMA3~\cite{zhang2025videollama3}, further improve video understanding through unified multimodal input, dynamic resolution processing, and long-context modeling, thereby enhancing visual perception and temporal reasoning.

Despite these advances, densely processing long videos remains difficult because the number of visual tokens grows rapidly with video duration. Therefore, recent studies have explored token compression and context window extension. For instance, SlowFast-LLaVA~\cite{xu2025slowfastllava15} adopts a slow-fast input design to reduce redundant visual tokens. LongVA~\cite{zhang2024longva} extends the context length of LLMs to accommodate more visual tokens. LongVU~\cite{shen2024longvu} introduces spatiotemporal adaptive compression to reduce redundancy while preserving visual details. Although these methods enhance the capacity of LVLMs to process longer videos, they do not eliminate the fundamental bottleneck: the number of raw frames in a long video still far exceeds what existing models can handle. Consequently, selecting informative frames remains a critical preprocessing step for long video understanding.

\subsection{Visual Sampling for Video Understanding}
Visual sampling aims to retain an informative subset of frames from long videos when LVLMs can only process a limited number of visual inputs. Existing methods can be broadly divided into training-based and training-free approaches.

\noindent\hspace{1em}\textbf{Training-based methods.}
Training-based methods learn additional modules or selection policies for visual sampling. For example, FFS~\cite{ffs} and Frame-Voyager~\cite{yu2025framevoyager} optimize visual sampling based on downstream task losses or frame combination ranking signals. Hu et al.~\cite{hu2025mllmselector} use advanced LVLMs to generate pseudo-labels for training a lightweight frame selector. Similarly, SeViLA~\cite{yu2023sevila} employs a vision-language model to jointly perform temporal localization and video question answering. Recent methods further explore more flexible learned samplers. GenS~\cite{yao2025gens} trains a generative frame sampler with dense frame-relevance annotations, while K-Frames~\cite{yao2025kframes} constructs frame-level annotations and combines supervised fine-tuning with reinforcement learning to support fine-grained retrieval and variable-size selection. Although these approaches learn effective selection policies, they rely heavily on dense annotations, pseudo-labels, model feedback, or task-specific training. In contrast, our framework eliminates the need for auxiliary training and functions as a plug-and-play preprocessing module for existing LVLMs.

\noindent\hspace{1em}\textbf{Training-free methods.}
Training-free methods select frames without updating model parameters, making them flexibly applicable to different LVLMs. Uniform sampling is a widely used training-free baseline that distributes the selected frames across the video, but it does not account for the visual evidence required by the query. Recent methods address this limitation by estimating the semantic relevance between the query and frames. For example, Q-Frame~\cite{zhang2025qframe} uses a text-image matching network to select query-relevant frames with multi-resolution adaptation. AKS~\cite{tang2025aks} jointly models query-frame relevance and temporal coverage, while KeyVideoLLM~\cite{liang2024keyvideollm} selects frames based on text-video frame similarity for efficient token compression. These methods improve the retrieval of query-relevant visual evidence, but high-relevance frames often concentrate around the same event, leading to redundant selections and insufficient visual coverage.

Recent studies further introduce diversity-aware selection. BOLT~\cite{liu2025bolt} improves sampling diversity through inverse transform sampling, while MDP$^3$~\cite{sun2025mdp3} formulates visual sampling as a Markov decision process that jointly optimizes query-frame relevance, list-level diversity, and temporal ordering. MaxInfo and AdaRD-Key select informative frames based on the maximum-volume principle and unified relevance-diversity objectives, respectively~\cite{li2025maxinfo,zhang2025adardkey}. EFS and WFS-SB adopt maximum marginal relevance objectives to trade-off relevance and diversity~\cite{chen2026efs,chen2026wfssb,carbonell1998mmr}. Overall, existing training-free visual sampling methods mainly improve temporal coverage, query-frame relevance, or diversity, and they do not explicitly adapt the selection process to different query requirements. Global queries require broad video coverage, whereas local queries require selecting frames from relevant event segments while maintaining coverage and reducing redundancy. This limitation motivates our visual sampling framework that preserves broad video context for global queries while concentrating the frame budget on relevant event segments for local queries.

\begin{figure*}[!t]
\centering
\includegraphics[width=1\textwidth]{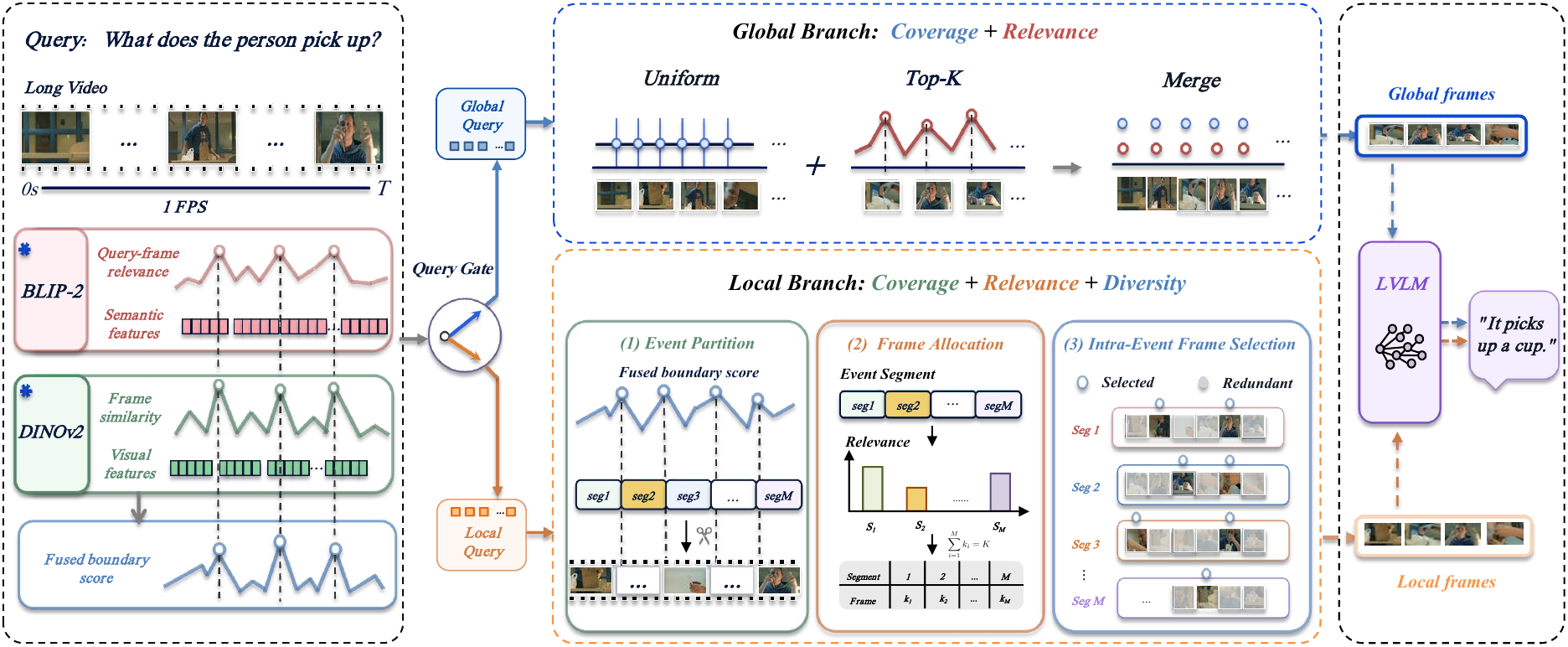}
\caption{Overview of VisualRouter. Each query is first routed to either the global or local branch. The global branch combines relevance-aware selection with temporal coverage, whereas the local branch performs event partitioning, segment-level frame allocation, and intra-segment frame selection.}
\label{fig:framework}
\end{figure*}

\section{Method}
In this section, we present VisualRouter, a training-free and plug-and-play framework for query-grounded visual sampling in long video understanding. As illustrated in Fig.~\ref{fig:framework}, given a video and query, VisualRouter first constructs a candidate frame sequence and routes the query to either the global or local branch according to its visual evidence requirements. The activated branch then computes the required visual signals and applies the corresponding sampling strategy to obtain the input frames. We begin with the problem formulation and then introduce each component of VisualRouter.

\subsection{Problem Formulation}
Given a long video $\mathbf{V}=\{I_t\}_{t=1}^{T}$ with $T$ frames and a textual query $Q$, visual sampling aims to retain an informative subset of frames that preserves the evidence required to answer $Q$. Let $K$ denote the maximum number of frames that can be processed by the LVLM, where $K \ll T$. Each selected frame is encoded into a sequence of visual tokens before being provided to the LLMs. If each frame contributes $m$ visual tokens and the textual prompt contains $\ell_Q$ tokens, the input must satisfy the constraint
\begin{equation}
K \cdot m + \ell_Q \leq L,
\end{equation}
where $L$ is the maximum context length of the LVLM.
Formally, the visual sampling problem can be expressed as
\begin{equation}
\mathcal{T}^{*} =
\arg\max_{\substack{\mathcal{T} \subseteq \{1,\ldots,T\},\ |\mathcal{T}|=K}}
\Phi(\mathcal{T}, \mathbf{V}, Q),
\end{equation}
where $\mathcal{T}=\{\tau_1,\ldots,\tau_K\}$ denotes the selected frame indices with $1 \leq \tau_1 < \tau_2 < \cdots < \tau_K \leq T$, and $\Phi$ evaluates the effectiveness of the selected frame subset in preserving the evidence required to answer $Q$.

The objective function $\Phi$ is designed to reflect three properties of the selected frames: query-frame relevance, evidence coverage, and visual diversity. Query-frame relevance favors frames containing visual content related to $Q$, while evidence coverage ensures that the selected subset captures sufficient evidence to answer the query. Visual diversity discourages redundant selections and encourages complementary information, particularly among frames in the same event segment.

However, the relative importance of these aspects varies across queries. Global queries require broad temporal coverage because their supporting evidence may be distributed throughout the video. In contrast, local queries require identifying relevant event segments, allocating sufficient frames to cover the supporting evidence, and reducing redundancy within each segment. Existing methods typically apply a fixed sampling policy to all queries, overlooking such differences in visual evidence requirements. VisualRouter addresses this limitation by first routing each query to either the global or local branch and then applying the corresponding sampling strategy.

\subsection{Visual Scoring and Feature Extraction}
\label{sec:signals}
To provide the visual signals required for visual sampling, we uniformly sample the input video at 1 FPS, obtaining a candidate frame sequence $\mathcal{I}=\{I_i\}_{i=1}^{T}$. For each candidate frame, we compute a query-frame relevance score and extract semantic and visual features. The relevance scores are used by both branches: the global branch uses them for relevance-aware selection, while the local branch uses them to estimate event relevance and allocate the frame. The semantic features support boundary detection and diversity-aware selection within event segments.

\noindent\hspace{1em}\textbf{Semantic Relevance Scores and Features.}
We use the image-text matching (ITM) head of the pretrained BLIP2 model~\cite{li2023blip2} to estimate the semantic relevance between each candidate frame and the textual input. Given a candidate frame $I_i$ and the textual query $\widetilde{Q}$, its relevance score is computed as
\begin{equation}
    s_i^{\mathrm{itm}} =
    \operatorname{\text{BLIP2-}ITM}(I_i,\widetilde{Q}),
    \quad i=1,\ldots,T .
\end{equation}
The score $s_i^{\mathrm{itm}}$ measures the semantic alignment between frame $I_i$ and the query $\widetilde{Q}$. We also extract the semantic representation $\mathbf{b}_i \in \mathbb{R}^{d_b}$ from the BLIP2 vision encoder and $\ell_2$-normalize it as $\hat{\mathbf{b}}_i=\mathbf{b}_i/\|\mathbf{b}_i\|_2$. The relevance scores $\{s_i^{\mathrm{itm}}\}_{i=1}^{T}$ support relevance-aware visual sampling and segment-level relevance estimation, whereas the normalized features $\{\hat{\mathbf{b}}_i\}_{i=1}^{T}$ are used to characterize semantic similarity and changes across candidate frames.

\noindent\hspace{1em}\textbf{Visual Features and Frame Similarity.}
Although BLIP2 features capture high-level semantic information, they may be less sensitive to low-level appearance changes, shot transitions, and changes in object layout. Therefore, we further extract visual features using DINOv2~\cite{oquab2024dinov2}. For each frame $I_i$, we extract a visual feature $\mathbf{d}_i \in \mathbb{R}^{d_d}$ and normalize it as $\hat{\mathbf{d}}_i=\mathbf{d}_i/\|\mathbf{d}_i\|_2$. Since DINOv2 provides strong self-supervised visual representations, these features are suitable for capturing appearance and scene changes. We compute the visual change between adjacent frames as
\begin{equation}
    s_i^{\mathrm{vis}} =
    1 - \langle \hat{\mathbf{d}}_i, \hat{\mathbf{d}}_{i-1} \rangle,
    \quad i=2,\ldots,T .
\end{equation}
The score $s_i^{\mathrm{vis}}$ measures the appearance change between consecutive candidate frames and serves as the visual signal for event partitioning in the local branch.

\subsection{Query Gating}
\label{sec:Query Gating}
Different questions require different visual sampling strategies. For instance, “\textit{What genre is this movie?}” requires an overall understanding of the video. In such cases, frame-level relevance scores may be noisy, and relevance-aware selection can be less effective than maintaining broad temporal coverage. VisualRouter therefore determines whether each query should be handled by the global or local branch before applying the corresponding sampling strategy.

We use the LVLM itself for query gating, without introducing an additional classifier or training. Given a question and its candidate, the LVLM predicts one of two routing labels. \textbf{LOCAL}: the question or at least one candidate contains a \emph{visual anchor}, defined as an element that can be directly recognized or localized in video frames, such as an object, scene, visible action, temporal cue, or OCR text. \textbf{GLOBAL}: both the question and all candidate are abstract, narrative, or evaluative and contain no clearly localizable visual anchor. We consider the question and answer options because an abstractly phrased question may still contain localizable visual cues in its options. The routing prompt is shown in Fig.~\ref{fig:prompt}.

For Video-MME and MLVU, the LVLM performs query gating independently for each question. For LongVideoBench (LVB), we directly route all samples to the local branch because its questions typically require locating and integrating visual evidence from one or more relevant temporal segments. This setting allows the local branch to explicitly model segment-level evidence relevance and coverage.

\begin{figure}[!t]
\centering
\begin{tikzpicture}
\node[
    draw=gray!35,
    fill=gray!3,
    rounded corners=2.5mm,
    line width=0.5pt,
    drop shadow={shadow xshift=1.5pt, shadow yshift=-1.5pt, opacity=0.25},
    inner xsep=5pt,
    inner ysep=6pt,
    text width=0.93\columnwidth,
    align=left
] (promptbox) {%
\begin{minipage}{0.90\columnwidth}
\vspace{2.5mm}
\fontsize{5.5pt}{6.0pt}\selectfont
\RaggedRight
\setlength{\parindent}{0pt}
\setlength{\parskip}{0pt}
You are an expert in video analysis. \\
Your task is to decide whether the question is better answered using local or global frame-level evidence.\\[0.4ex]

\textbf{Routing Labels}\\[-0.3ex]
\begin{itemize}[leftmargin=1.2em, itemsep=0pt, topsep=0pt, parsep=0pt]
    \item \texttt{LOCAL}: The question or at least one candidate option contains a concrete visual anchor that allows the supporting
    evidence to be localized to one or more video segments.
    \item \texttt{GLOBAL}: Neither the question nor its candidate options contain a reliable visual anchor, and answering the question
    instead requires broad video-level context.
\end{itemize}

\textbf{Visual Anchor Definition}\\[-0.2ex]
A visual anchor is a directly observable or localizable element, such as
a specific person, object, scene, visible action, displayed text, or
temporally localized event. Abstract attributes, narrative interpretations,
and video-level categories are not visual anchors unless they can be
reliably grounded in particular frames or segments.\\[0.4ex]

\textbf{Decision Rules}\\[-0.3ex]
\begin{enumerate}[leftmargin=1.35em, itemsep=0pt, topsep=0pt,
parsep=0pt, label=\arabic*.]
    \item Examine both the question and all candidate answer options.
    \item If either contains a reliable visual anchor, output
    \texttt{LOCAL}.
    \item Otherwise, output \texttt{GLOBAL}.
\end{enumerate}

\textbf{Examples}\\[-0.2ex]
\textit{Local example.}
Q: ``What is the primary focus of the video?''
The options include concrete activities such as ``fitness exercises''
and ``eating and shopping''.
$\Rightarrow$ \texttt{LOCAL}.\\[-0.1ex]

\textit{Global example.}
Q: ``What is the plot of the opera?''
All options are narrative summaries without directly localizable visual
anchors.
$\Rightarrow$ \texttt{GLOBAL}.\\[0.4ex]

\textbf{\textit{Input}}\\[-0.2ex]
\textbf{Question:}~\textit{\{question text\}}\\
\textbf{Candidate Options:}~
A.~\textit{\{opt A\}};
B.~\textit{\{opt B\}};
C.~\textit{\{opt C\}};
D.~\textit{\{opt D\}}\\[0.4ex]

\textbf{\textit{Output}}\\[-0.2ex]
\texttt{\{"label": "local"\}} \quad or \quad
\texttt{\{"label": "global"\}}

\end{minipage}
};

\node[
    anchor=west,
    fill=gray!55,
    draw=gray!45,
    rounded corners=1mm,
    line width=0.4pt,
    drop shadow={shadow xshift=0.8pt, shadow yshift=-0.8pt, opacity=0.25},
    inner xsep=4pt,
    inner ysep=1.5pt,
    font=\bfseries\fontsize{6.5pt}{7.2pt}\selectfont
] at ([xshift=7pt,yshift=-1pt]promptbox.north west)
{System Prompt: Query Gating};

\end{tikzpicture}
\caption{Prompt design for query gating. The prompt determines whether a question requires local frame-level evidence or global video-level evidence by detecting concrete visual anchors in both the question and candidate options.}
\label{fig:prompt}
\end{figure}

\subsection{Query-Grounded Visual Sampling}
\label{sec:Query-Routed Frame Selection}
Based on the routing result from Sec.~\ref{sec:Query Gating}, VisualRouter activates either the global or local selection branch, as illustrated in Fig.~\ref{fig:framework}. Both branches operate on the candidate frame sequence $\mathcal{I}$ and select $K$ frames using their corresponding visual signals defined in Sec.~\ref{sec:signals}. The global branch adopts a lightweight strategy that combines temporal coverage with query relevance, whereas the local branch follows a three-stage event-aware selection pipeline. Each branch is detailed in the following subsections.

\subsubsection{Global-Query Relevance-Coverage Sampling}
For global queries, selecting frames solely according to query-frame relevance may fail to preserve sufficient video-level context. Because such queries often lack explicit visual anchors, the relevance estimates may be noisy or concentrated on only a few temporally adjacent frames. In contrast, uniform sampling provides broad temporal coverage but overlooks query-relevant visual evidence. We therefore combine temporally uniform sampling with relevance-aware frame selection.

Specifically, we allocate $\lfloor K/2 \rfloor$ frames to temporal coverage and sample them uniformly from the candidate sequence $\mathcal{I}$. The remaining $K-\lfloor K/2 \rfloor$ frames are selected from the unselected candidates with the highest query-frame relevance scores $\{s_i^{\mathrm{itm}}\}_{i=1}^{T}$ defined in Sec.~\ref{sec:signals}. The two subsets are merged and sorted in temporal order to obtain the selected frame set for global queries. If duplicate frames appear in the two subsets, we remove them and fill the remaining slots with the highest-scoring unselected frames. This strategy preserves broad video context while retaining query-relevant evidence, without additional training or model-specific modifications.

\subsubsection{Local-Query Event-Aware Sampling}
\label{sec:event_selection}
For local queries, the required visual evidence is typically concentrated in one or a few relevant event segments. Uniform sampling may allocate selected frames to irrelevant segments, whereas query-frame relevance selection often retrieves temporally adjacent frames with highly similar content, resulting in redundant evidence and incomplete coverage of the relevant event. 
To address these limitations, we introduce a three-stage event-aware selection pipeline: (i) \emph{event partitioning}, which partitions the candidate frame sequence into event segments; (ii) \emph{segment-level frame allocation}, which estimates the relevance of each event and distributes the frame budget accordingly; and (iii) \emph{intra-event frame selection}, which selects query-relevant and non-redundant frames within each allocated event segment. Implementation details are provided in Algorithm~\ref{alg:event_anchored}.

\noindent\hspace{1em}\textbf{Event Partitioning.}
Direct comparisons between adjacent frame features can be sensitive to short-term noise, causing minor frame-level fluctuations to be mistaken for event transitions. To obtain a more stable semantic reference, we maintain an exponential moving average (EMA) of the normalized BLIP2 features. We initialize $\boldsymbol{\mu}_1=\hat{\mathbf{b}}_1$ and update the reference as
\begin{equation}
    \boldsymbol{\mu}_i
    =
    (1-\rho)\boldsymbol{\mu}_{i-1}
    +
    \rho\,\hat{\mathbf{b}}_i,
    \quad i=2,\ldots,T.
\end{equation}
where $\rho$ controls the update rate. We define semantic drift as the deviation of the current frame from an EMA representation of the preceding video context
\begin{equation}
    s_i^{\mathrm{sem}}
    =
    1-\left\langle
    \hat{\mathbf{b}}_i,
    \boldsymbol{\mu}_{i-1}
    \right\rangle,
    \quad i=2,\ldots,T.
\end{equation}
For each frame, we first compute $s_i^{\mathrm{sem}}$ by comparing its feature $\hat{\mathbf{b}}_i$ with the preceding EMA reference $\boldsymbol{\mu}_{i-1}$. We then incorporate $\hat{\mathbf{b}}_i$ into the EMA to obtain $\boldsymbol{\mu}_i$. A larger semantic drift score indicates a greater semantic difference from the recent video context and may correspond to an event transition.

In addition to semantic drift, we use the visual change score $s_i^{\mathrm{vis}}$ defined in Sec.~\ref{sec:signals}. It measures the cosine distance between consecutive DINOv2 features and captures abrupt appearance changes and shot transitions, providing a complementary cue for event partitioning. We apply smoothing and robust normalization before fusion
\begin{equation}
    p_i
    =
    \lambda_s\,\widetilde{s}_i^{\mathrm{sem}}
    +
    \lambda_v\,\widetilde{s}_i^{\mathrm{vis}}.
\end{equation}
where $\lambda_s$ and $\lambda_v$ are weights for the semantic and visual boundary signals, respectively. We then apply local peak detection to the fused boundary score $\{p_i\}_{i=2}^{T}$ to identify event boundaries. The detected boundaries partition the candidate frame sequence into $M$ consecutive event segments $\mathcal{C}=\{C_m\}_{m=1}^{M}$, where each segment is denoted as $C_m=[a_m,b_m)$.

\begingroup
\renewcommand{\algorithmicrequire}{\textbf{Input:}}
\renewcommand{\algorithmicensure}{\textbf{Output:}}
\begin{algorithm}[t]
\caption{Event-Aware Frame Selection}
\label{alg:event_anchored}
\begin{algorithmic}[1]

\REQUIRE Candidate frames $\mathcal{I}$, frame budget $K$,
relevance scores $\mathbf{s}^{\mathrm{itm}}$,
BLIP2 features $\hat{\mathbf{B}}$,
and DINOv2 features $\hat{\mathbf{D}}$.
\ENSURE Selected frame set $\mathcal{T}$.

\STATE \textbf{Stage 1: Event Partitioning.}
\STATE Compute $\mathbf{s}^{\mathrm{sem}}$ from $\hat{\mathbf{B}}$
and $\mathbf{s}^{\mathrm{vis}}$ from $\hat{\mathbf{D}}$.
\STATE Smooth and normalize to obtain
$\widetilde{\mathbf{s}}^{\mathrm{sem}}$ and
$\widetilde{\mathbf{s}}^{\mathrm{vis}}$.
\STATE $p_i \gets
\lambda_s\widetilde{s}_i^{\mathrm{sem}}
+\lambda_v\widetilde{s}_i^{\mathrm{vis}}$.
\STATE Detect boundary peaks from $\mathbf{p}$ and partition
$\mathcal{I}$ into event segments
$\mathcal{C}=\{C_m\}_{m=1}^{M}$.

\STATE \textbf{Stage 2: Segment-Level Frame Allocation.}
\FOR{each segment $C_m=[a_m,b_m)$}
    \STATE Compute $\ell_m$, $s_m^{\max}$, and $s_m^{\mathrm{mean}}$.
    \STATE $w_m\gets
    \sqrt{s_m^{\max}s_m^{\mathrm{mean}}}\cdot\sqrt{\ell_m}$.
\ENDFOR
\STATE Allocate frame counts $\{k_m\}_{m=1}^{M}$
from $\{w_m\}_{m=1}^{M}$.

\STATE \textbf{Stage 3: Intra-Event Frame Selection.}
\STATE $\mathcal{T}\gets\emptyset$.
\FOR{each segment $C_m$ with $k_m>0$}
    \STATE $q_i\gets (s_i^{\mathrm{itm}})^\gamma$, \quad
    $L^{(m)}_{ij}\gets
    q_i\langle\hat{\mathbf{b}}_i,\hat{\mathbf{b}}_j\rangle q_j$.
    \STATE $S_m\gets
    \operatorname{GreedyDPP}(C_m,k_m,L^{(m)})$.
    \STATE $\mathcal{T}\gets\mathcal{T}\cup S_m$.
\ENDFOR
\RETURN $\operatorname{Sort}(\mathcal{T})$.

\end{algorithmic}
\end{algorithm}

\noindent\hspace{1em}\textbf{Segment-Level Frame Allocation.}
After partitioning the video into event segments, we allocate the $K$ frames across them. For local queries, the relevant visual evidence is usually concentrated in one or a few event segments, so segments with higher query-frame relevance should receive more frames.
Segment length is also considered because longer events may contain richer temporal variations and require broader coverage. We thus determine the allocation jointly from segment relevance and duration.

Given the event segments $\mathcal{C}=\{C_m\}_{m=1}^{M}$, let $C_m=[a_m,b_m)$ denote the $m$-th segment, with length $\ell_m=b_m-a_m$. We characterize its relevance using the peak and mean frame-level scores:
\begin{equation}
    s_m^{\max}
    =
    \max\left(
        0,
        \max_{i\in C_m}s_i^{\mathrm{itm}}
    \right).
\end{equation}

\begin{equation}
    s_m^{\mathrm{mean}}
    =
    \max\left(
        0,
        \frac{1}{\ell_m}
        \sum_{i\in C_m}s_i^{\mathrm{itm}}
    \right).
\end{equation}
Here, $s_m^{\max}$ indicates whether the segment contains highly query-relevant frames, while $s_m^{\mathrm{mean}}$ provides a stable estimate of the overall segment-level relevance. The peak score captures strong localized evidence, whereas the mean score provides a more robust estimate of the overall relevance of the segment. We combine the two scores with segment length to define the importance weight
\begin{equation}
    w_m
    =
    \sqrt{s_m^{\max}\cdot s_m^{\mathrm{mean}}}
    \cdot
    \sqrt{\ell_m}.
\end{equation}
This weight reflects both query-frame relevance and the need for temporal coverage. When $\sum_{j=1}^{M}w_j>0$, the allocation for segment $C_m$ is computed as:
\begin{equation}
    \hat{k}_m
    =
    K\cdot
    \frac{w_m}{\sum_{j=1}^{M}w_j}.
\end{equation}
The final number of frames $k_m$ assigned to each segment is obtained by taking the floor of $\hat{k}_m$ and assigning the remaining frames to the segments with the largest fractional parts, ensuring $\sum_{m=1}^{M}k_m=K$. If all segment weights are zero, we fall back to uniform or length-based allocation to ensure a valid frame allocation.

\noindent\hspace{1em}\textbf{Intra-Event Frame Selection.}
After allocating $k_m$ frames to each event segment $C_m$, we select the corresponding frames within each segment. Directly selecting the Top-$k_m$ frames by query-frame relevance often retrieves temporally adjacent frames with highly similar semantic content, resulting in redundant visual evidence. We therefore select frames by jointly considering query-frame relevance and inter-frame diversity, so that each segment contributes informative and non-redundant visual evidence.

We instantiate a query-aware $k$-DPP within each event segment~\cite{kulesza2012dpp}, which favors subsets containing high-quality yet mutually dissimilar items. In our setting, frame quality is determined by query-frame relevance, while frame similarity is computed from BLIP2 semantic features. Because the number of frames assigned to segment $C_m$ is fixed as $k_m$, we perform size-$k_m$ DPP selection within the segment~\cite{kulesza2011kdpp}.

Specifically, for each candidate frame $I_i$ in segment $C_m$, we define its quality score as $q_i=f_q(s_i^{\mathrm{itm}})$, where $s_i^{\mathrm{itm}}$ is the query-frame relevance score computed by BLIP2 ITM. We use $f_q(s)=s^\gamma$ with $\gamma>1$ to increase the relative weight of highly relevant frames while suppressing less relevant frames. The semantic similarity between frames $I_i$ and $I_j$ is measured by the inner product of their normalized BLIP2 semantic features, i.e., $A_{ij}=\langle\hat{\mathbf{b}}_i,\hat{\mathbf{b}}_j\rangle$. The DPP kernel for segment $C_m$ is constructed as
\begin{equation}
   L_{ij}^{(m)} = q_i A_{ij} q_j,
\qquad i,j\in C_m. 
\end{equation}
Given the kernel $L^{(m)}$, we select exactly $k_m$ frames by maximizing the determinant:
\begin{equation}
    S_m^*
    =
    \arg\max_{\substack{S\subseteq C_m\\ |S|=k_m}}
    \det\left(L_S^{(m)}\right).
\end{equation}
Here, $L^{(m)}_S$ denotes the principal submatrix of $L^{(m)}$ indexed by $S$. The determinant jointly reflects frame quality and diversity. If two selected frames are highly similar, the corresponding rows and columns of $L^{(m)}_S$ become nearly dependent, resulting in a smaller determinant. In contrast, a subset containing query-relevant and semantically complementary frames receives a higher determinant score. Since exact subset search is computationally expensive, we use the standard greedy approximation for log-determinant maximization~\cite{chen2018fast}. Starting from $S_m=\emptyset$, we repeatedly add the frame with the largest marginal log-determinant gain:
\begin{equation}
    t^*
    =
    \operatorname*{arg\,max}_{
        \substack{
            t\in C_m \\
            t\notin S_m
        }
    }
    \left[
        \log\det\bigl(L^{(m)}_{S_m\cup\{t\}}\bigr)
        -
        \log\det\bigl(L^{(m)}_{S_m}\bigr)
    \right].
\end{equation}
We add $t^*$ to $S_m$ and repeat this process until $k_m$ frames are selected. Finally, the selected subsets from all event segments are merged and sorted in temporal order to obtain the final frame set $\mathcal{T}$, which is concatenated with the query $Q$ and fed into the pretrained LVLM.

\section{Experiments}
\subsection{Experiment Settings}
\subsubsection{Evaluation Benchmarks}
To comprehensively evaluate the effectiveness of our proposed method on long video understanding tasks, we conduct experiments on three representative benchmarks: Video-MME~\cite{fu2025videomme}, LongVideoBench~\cite{wu2024longvideobench}, and MLVU~\cite{zhou2025mlvu}. These benchmarks cover diverse video durations and task types, enabling a systematic assessment of the generalization and effectiveness of visual sampling strategies. 

Video-MME consists of 900 videos and 2,700 human-annotated question-answer pairs, with video durations ranging from 11 seconds to 60 minutes and averaging approximately 17 minutes. It provides a balanced distribution over short videos ($<$2 minutes), medium videos (4-15 minutes), and long videos (30-60 minutes), making it suitable for evaluating model understanding at different temporal scales. LongVideoBench contains 1,337 validation QA pairs with an average video duration of about 12 minutes, focusing on long video temporal reasoning and cross-segment information integration. MLVU is a multi-task long video understanding benchmark comprising 2,174 question-answer pairs, designed to assess event understanding, action recognition, and multi-level temporal understanding.

\subsubsection{Evaluation Framework and Models}
All experiments are conducted using the LMMs-Eval~\cite{zhang2025lmmseval} framework to ensure reproducible evaluation. To verify the generality of VisualRouter, we integrate it with multiple representative LVLMs as a plug-and-play visual sampling module, including LLaVA-Video-7B~\cite{zhang2025llavavideo}, LLaVA-OneVision-7B~\cite{li2024llavaonevision}, Qwen2.5-VL-7B~\cite{bai2025qwen25vl}, Qwen3-VL-8B~\cite{bai2025qwen3vl}, and InternVL3-8B~\cite{zhu2025internvl3}. For fair comparison, all methods use the same LVLM backbone, input prompts, and number of input frames. No subtitles are used, and only the sampling strategy is changed.

\begin{figure*}[!t]
\centering
\includegraphics[width=0.85\textwidth]{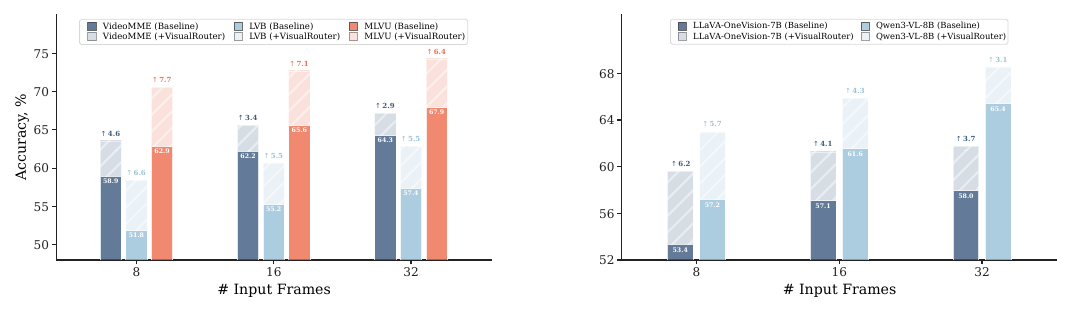}
\caption{Effect of the number of input frames. Dark bars show the accuracy of uniform sampling, while light extensions show the accuracy gains achieved by VisualRouter. (a) InternVL3-8B on three benchmarks. (b) Multiple LVLMs on Video-MME.}
\label{fig:frame_budget}
\end{figure*}

\begin{table*}[!t]
\caption{Comprehensive evaluation on Video-MME, LongVideoBench (LVB), and MLVU. Representative long-video LVLMs are included as reference results because they may use different backbones, training data, input frames, and input strategies. The lower block evaluates VisualRouter as a plug-and-play visual sampling module under the same LVLM and the same number of input frames. $\ddagger$ indicates frames selected by VisualRouter.}
\label{tab:lvlm_comparison}
\centering
\footnotesize
\setlength{\tabcolsep}{3pt}
\renewcommand{\arraystretch}{1.15}

\begin{tabular}{lcccccccc}
\toprule
\multirow{2}{*}{\textbf{Model}} &
\multirow{2}{*}{\textbf{LLM Size}} &
\multirow{2}{*}{\textbf{\#Frames}} &
\multicolumn{4}{c}{\textbf{Video-MME (w.o. sub.)}} &
\multirow{2}{*}{\textbf{LVB}} &
\multirow{2}{*}{\textbf{MLVU}} \\
\cmidrule(lr){4-7}
& & &
\textbf{Short} &
\textbf{Medium} &
\textbf{Long} &
\textbf{Overall} &
& \\
\midrule

Video-LLaVA~\cite{lin2023videollava}
& 7B & 8
& 45.3 & 38.0 & 36.2 & 39.9 & 39.1 & 47.3 \\

LongVILA~\cite{chen2024longvila}
& 8B & 128
& 60.2 & 48.2 & 38.8 & 49.2 & -- & -- \\

Video-XL~\cite{shu2025videoxl}
& 7B & 128
& 64.0 & 53.2 & 49.2 & 55.5 & -- & 64.9 \\

LongVU~\cite{shen2024longvu}
& 7B & 1 FPS
& 64.7 & 58.2 & 59.5 & 60.9 & -- & 65.4 \\

\midrule

LLaVA-OneVision~\cite{li2024llavaonevision}
& 7B & 16
& 69.0 & 54.4 & 48.0 & 57.1 & 55.0 & 60.5 \\

\quad\textbf{+ VisualRouter}
& 7B & 16$^\ddagger$
& \textbf{72.0} {\scriptsize\textcolor{violet!80!black}{(+3.0)}}
& \textbf{61.0} {\scriptsize\textcolor{violet!80!black}{(+6.6)}}
& \textbf{50.8} {\scriptsize\textcolor{violet!80!black}{(+2.8)}}
& \textbf{61.3} {\scriptsize\textcolor{violet!80!black}{(+4.2)}}
& \textbf{62.3} {\scriptsize\textcolor{violet!80!black}{(+7.3)}}
& \textbf{67.7} {\scriptsize\textcolor{violet!80!black}{(+7.2)}} \\

\midrule

Qwen2.5-VL~\cite{bai2025qwen25vl}
& 7B & 16
& 67.3 & 55.0 & 48.9 & 57.1 & 56.0 & 56.3 \\

\quad\textbf{+ VisualRouter}
& 7B & 16$^\ddagger$
& \textbf{73.3} {\scriptsize\textcolor{violet!80!black}{(+6.0)}}
& \textbf{62.0} {\scriptsize\textcolor{violet!80!black}{(+7.0)}}
& \textbf{51.7} {\scriptsize\textcolor{violet!80!black}{(+2.8)}}
& \textbf{62.3} {\scriptsize\textcolor{violet!80!black}{(+5.2)}}
& \textbf{63.7} {\scriptsize\textcolor{violet!80!black}{(+7.7)}}
& \textbf{67.9} {\scriptsize\textcolor{violet!80!black}{(+11.6)}} \\

\midrule

LLaVA-Video~\cite{zhang2025llavavideo}
& 7B & 32
& 75.7 & 59.6 & 52.9 & 62.7 & 58.0 & 64.0 \\

\quad\textbf{+ VisualRouter}
& 7B & 32$^\ddagger$
& \textbf{76.6} {\scriptsize\textcolor{violet!80!black}{(+0.9)}}
& \textbf{64.8} {\scriptsize\textcolor{violet!80!black}{(+5.2)}}
& \textbf{54.4} {\scriptsize\textcolor{violet!80!black}{(+1.5)}}
& \textbf{65.2} {\scriptsize\textcolor{violet!80!black}{(+2.5)}}
& \textbf{63.9} {\scriptsize\textcolor{violet!80!black}{(+5.9)}}
& \textbf{69.4} {\scriptsize\textcolor{violet!80!black}{(+5.4)}} \\

\midrule

Qwen3-VL~\cite{bai2025qwen3vl}
& 8B & 32
& 77.9 & 62.3 & 56.0 & 65.4 & 57.4 & 62.7 \\

\quad\textbf{+ VisualRouter}
& 8B & 32$^\ddagger$
& \textbf{77.9} {\scriptsize\textcolor{violet!80!black}{(+0.0)}}
& \textbf{67.9} {\scriptsize\textcolor{violet!80!black}{(+5.6)}}
& \textbf{59.8} {\scriptsize\textcolor{violet!80!black}{(+3.8)}}
& \textbf{68.5} {\scriptsize\textcolor{violet!80!black}{(+3.1)}}
& \textbf{64.4} {\scriptsize\textcolor{violet!80!black}{(+7.0)}}
& \textbf{74.9} {\scriptsize\textcolor{violet!80!black}{(+12.2)}} \\

\midrule

InternVL3-8B~\cite{zhu2025internvl3}
& 8B & 32
& 75.8 & 63.6 & 53.4 & 64.3 & 57.4 & 67.9 \\

\quad\textbf{+ VisualRouter}
& 8B & 32$^\ddagger$
& \textbf{76.1} {\scriptsize\textcolor{violet!80!black}{(+0.3)}}
& \textbf{68.1} {\scriptsize\textcolor{violet!80!black}{(+4.5)}}
& \textbf{57.2} {\scriptsize\textcolor{violet!80!black}{(+3.8)}}
& \textbf{67.1} {\scriptsize\textcolor{violet!80!black}{(+2.8)}}
& \textbf{62.8} {\scriptsize\textcolor{violet!80!black}{(+5.4)}}
& \textbf{74.3} {\scriptsize\textcolor{violet!80!black}{(+6.4)}} \\

\bottomrule
\end{tabular}
\end{table*}

\subsubsection{Implementation Details}
To reduce computational cost, we sample each video at 1 FPS to construct a candidate frame sequence. Query-frame relevance scores are computed using the BLIP2-ITM~\cite{li2023blip2}, while DINOv2~\cite{oquab2024dinov2} is used for visual feature extraction. For event partitioning, we set the fusion weights to $\lambda_s=0.6$ and $\lambda_v=0.4$ for the semantic and visual signals, respectively.  Given the number of input frames $K$, VisualRouter selects $K$ frames and feeds them into the LVLM as visual inputs. The uniform baseline selects $K$ frames, ensuring that all methods use the same number of visual inputs. All experiments follow the official evaluation protocols of the corresponding benchmarks.

\begin{table*}[!t]
\caption{Comparison with representative training-free visual sampling methods
on Video-MME, LongVideoBench (LVB), and MLVU. Within each model group,
all methods use the same LVLM and number of input frames.
$\dagger$ denotes reproduced results.}
\label{tab:frame_method_comparison}
\centering
\footnotesize
\begin{tabular}{llcccccccc}
\toprule
\multirow{2}{*}{\textbf{Model}} &
\multirow{2}{*}{\textbf{Method}} &
\multirow{2}{*}{\textbf{LLM Size}} &
\multirow{2}{*}{\textbf{\#Frames }} &
\multicolumn{4}{c}{\textbf{Video-MME (w.o. sub)}} &
\multirow{2}{*}{\textbf{LVB}} &
\multirow{2}{*}{\textbf{MLVU}} \\
\cmidrule(lr){5-8}
& & & &
\textbf{Short} &
\textbf{Medium} &
\textbf{Long} &
\textbf{Overall} &
& \\
\midrule

\multirow{6}{*}{Qwen2.5-VL}
& Uniform             & 7B & 16 & 67.3 & 55.0 & 48.9 & 57.1 & 56.0 & 56.3 \\
& Top-$K$             & 7B & 16 & 71.4 & 60.4 & 50.7 & 60.9 & 61.5 & 66.0 \\
& BOLT$^\dagger$~\cite{liu2025bolt} & 7B & 16 & 71.1 & 58.6 & 51.4 & 60.4 & 58.3 & 63.6 \\
& AKS$^\dagger$~\cite{tang2025aks}  & 7B & 16 & 71.2 & 59.6 & 51.7 & 60.8 & 59.9 & 64.3 \\
& WFS-SB$^\dagger$~\cite{chen2026wfssb} & 7B & 16 & 72.2 & 61.8 & 50.8 & 61.6 & 62.1 & 67.6 \\
& \textbf{VisualRouter}  & 7B & 16
& 73.3
& 62.0
& 51.7
& \textbf{62.3} {\scriptsize\textcolor{violet!80!black}{(+5.2)}}
& \textbf{63.7} {\scriptsize\textcolor{violet!80!black}{(+7.7)}}
& \textbf{67.9} {\scriptsize\textcolor{violet!80!black}{(+11.6)}} \\
\midrule

\multirow{6}{*}{LLaVA-OneVision}
& Uniform             & 7B & 16 & 69.0 & 54.4 & 48.0 & 57.1 & 55.0 & 60.5 \\
& Top-$K$             & 7B & 16 & 71.3 & 60.2 & 46.8 & 59.4 & 61.2 & 64.4 \\
& BOLT$^\dagger$~\cite{liu2025bolt} & 7B & 16 & 68.2 & 57.7 & 49.2 & 58.4 & 56.6 & 64.3 \\
& AKS$^\dagger$~\cite{tang2025aks}  & 7B & 16 & 69.7 & 56.8 & 49.7 & 58.7 & 59.4 & 67.0 \\
& WFS-SB$^\dagger$~\cite{chen2026wfssb} & 7B & 16 & 72.2 & 62.1 & 48.2 & 60.9 & 61.6 & \textbf{68.0} \\
& \textbf{VisualRouter}  & 7B & 16
& 72.0
& 61.0
& 50.8
& \textbf{61.3} {\scriptsize\textcolor{violet!80!black}{(+4.2)}}
& \textbf{62.3} {\scriptsize\textcolor{violet!80!black}{(+7.3)}}
& 67.7 {\scriptsize\textcolor{violet!80!black}{(+7.2)}} \\
\midrule

\multirow{6}{*}{LLaVA-Video}
& Uniform             & 7B & 32 & 75.7 & 59.6 & 52.9 & 62.7 & 58.0 & 64.0 \\
& Top-$K$             & 7B & 32 & 77.1 & 61.8 & 51.4 & 63.4 & 62.2 & 67.2 \\
& BOLT$^\dagger$~\cite{liu2025bolt} & 7B & 32 & 75.3 & 63.0 & 56.2 & 64.9 & 60.2 & 67.4 \\
& AKS$^\dagger$~\cite{tang2025aks}  & 7B & 32 & 76.3 & 63.2 & 53.9 & 64.5 & 60.9 & 67.9 \\
& WFS-SB$^\dagger$~\cite{chen2026wfssb} & 7B & 32 & 77.3 & 64.6 & 53.3 & 65.1 & 62.6 & 69.3 \\
& \textbf{VisualRouter}  & 7B & 32
& 76.6
& 64.8
& 54.4
& \textbf{65.2} {\scriptsize\textcolor{violet!80!black}{(+2.5)}}
& \textbf{63.9} {\scriptsize\textcolor{violet!80!black}{(+5.9)}}
& \textbf{69.4} {\scriptsize\textcolor{violet!80!black}{(+5.4)}} \\
\midrule

\multirow{6}{*}{Qwen3-VL}
& Uniform             & 8B & 32 & 77.9 & 62.3 & 56.0 & 65.4 & 57.4 & 62.7 \\
& Top-$K$             & 8B & 32 & 79.3 & 64.2 & 57.8 & 67.1 & 64.1 & 73.3 \\
& BOLT$^\dagger$~\cite{liu2025bolt} & 8B & 32 & 76.2 & 65.4 & 58.0 & 66.6 & 59.9 & 69.1 \\
& AKS$^\dagger$~\cite{tang2025aks}  & 8B & 32 & 78.2 & 64.4 & 56.6 & 66.4 & 60.1 & 70.4 \\
& WFS-SB$^\dagger$~\cite{chen2026wfssb} & 8B & 32 & 79.7 & 65.8 & 57.7 & 67.7 & \textbf{65.0} & 73.1 \\
& \textbf{VisualRouter}  & 8B & 32
& 77.9
& 67.9
& 59.8
& \textbf{68.5} {\scriptsize\textcolor{violet!80!black}{(+3.1)}}
& 64.4 {\scriptsize\textcolor{violet!80!black}{(+7.0)}}
& \textbf{74.9} {\scriptsize\textcolor{violet!80!black}{(+12.2)}} \\
\midrule

\multirow{6}{*}{InternVL3}
& Uniform             & 8B & 32 & 75.8 & 63.6 & 53.4 & 64.3 & 57.4 & 67.9 \\
& Top-$K$             & 8B & 32 & 77.0 & 64.3 & 53.9 & 65.1 & 62.7 & 72.7 \\
& BOLT$^\dagger$~\cite{liu2025bolt} & 8B & 32 & 76.1 & 66.8 & 56.0 & 66.3 & 59.6 & 70.4 \\
& AKS$^\dagger$~\cite{tang2025aks}  & 8B & 32 & 77.3 & 68.3 & 56.7 & \textbf{67.4} & 60.0 & 74.0 \\
& WFS-SB$^\dagger$~\cite{chen2026wfssb} & 8B & 32 & 77.0 & 67.1 & 54.7 & 66.3 & 61.3 & 73.9 \\
& \textbf{VisualRouter}  & 8B & 32
& 76.1
& 68.1
& 57.2
& 67.1 {\scriptsize\textcolor{violet!80!black}{(+2.8)}}
& \textbf{62.8} {\scriptsize\textcolor{violet!80!black}{(+5.4)}}
& \textbf{74.3} {\scriptsize\textcolor{violet!80!black}{(+6.4)}} \\
\bottomrule
\end{tabular}
\end{table*}

\subsection{Comparison with Existing Methods}
\subsubsection{Performance with Different LVLMs}
To evaluate the generality of VisualRouter, we integrate it with different LVLM backbones and compare it with the default uniform sampling strategy under the same number of input frames. We keep the model parameters, prompts, and evaluation settings fixed and vary only the visual sampling strategy. As reported in Table~\ref{tab:lvlm_comparison}, VisualRouter consistently improves the performance of different LVLM backbones across Video-MME, LongVideoBench, and MLVU, with accuracy gains ranging from 2.5\% to 12.2\%. This suggests that VisualRouter improves long video understanding by adaptively selecting informative frames and preserving critical visual and temporal evidence. These results show that VisualRouter generalizes well across different LVLM models.

\subsubsection{Performance of Different Visual Sampling Methods}
We compare VisualRouter with representative training-free visual sampling methods, including Top-$K$, BOLT, AKS, and WFS-SB, under the same LVLM backbone and number of input frames. As shown in Table~\ref{tab:frame_method_comparison}, VisualRouter consistently outperforms uniform sampling across all backbones and benchmarks. Across different video durations, VisualRouter achieves more pronounced improvements on medium and long videos, while remaining comparable to other sampling methods on short videos. This indicates that adaptive visual sampling is particularly beneficial when the video contains longer temporal contexts and more complex evidence distributions. For example, with LLaVA-Video-7B using 32 frames, it improves accuracy by 2.5, 5.9, and 5.4 percentage points on Video-MME, LongVideoBench, and MLVU, respectively. Moreover, VisualRouter achieves the highest or competitive accuracy across the 15 benchmark settings. These results suggest that routing queries to different visual sampling strategies according to their evidence requirements is more effective than applying a single fixed strategy based solely on relevance, coverage, or diversity.

Fig.~\ref{fig:question_type} further compares different visual sampling methods across six question categories on Video-MME. VisualRouter consistently outperforms uniform sampling in all categories and achieves the best performance in five out of six. Notably, the largest gains are observed on OCR and counting, with improvements of 12.95\% and 7.09\%, respectively, highlighting its ability to identify sparse and query-relevant visual evidence.

\subsection{Ablation and Analysis}
\subsubsection{Effect of the Number of Frames}
We evaluate VisualRouter with different numbers of frames in Fig.~\ref{fig:frame_budget}. Figure~\ref{fig:frame_budget}(a) reports the results of InternVL3-8B across three benchmarks, while Figure~\ref{fig:frame_budget}(b) compares different LVLMs on Video-MME. VisualRouter outperforms uniform sampling across all frame budgets, with particularly large gains under the 8-frame setting, ranging from 4.6 to 7.7 percentage points across the three benchmarks. This indicates that VisualRouter makes more effective use of a limited frame budget by prioritizing frames that are most useful for answering the query.

\begin{table}[!t]
\caption{Evaluation of VisualRouter across Qwen2.5-VL model scales
with $K=16$ input frames. Accuracy (\%) is reported, with improvements over
uniform sampling shown in parentheses.}

\label{tab:scale_ablation}
\centering
\footnotesize
\begin{tabular}{clcc}
\toprule
Benchmark & LVLM Scale & Uniform & VisualRouter \\
\midrule
\multirow{4}{*}{Video-MME}
& Qwen2.5-VL-3B  & 54.3 & 59.6 {\scriptsize\textcolor{violet!80!black}{(+5.3)}} \\
& Qwen2.5-VL-7B  & 57.1 & 62.3 {\scriptsize\textcolor{violet!80!black}{(+5.2)}} \\
& Qwen2.5-VL-32B & 59.9 & 64.2 {\scriptsize\textcolor{violet!80!black}{(+4.3)}} \\
& Qwen2.5-VL-72B & 63.2 & 67.1 {\scriptsize\textcolor{violet!80!black}{(+3.9)}} \\
\midrule
\multirow{4}{*}{LVB}
& Qwen2.5-VL-3B  & 54.1 & 57.8 {\scriptsize\textcolor{violet!80!black}{(+3.7)}} \\
& Qwen2.5-VL-7B  & 56.0 & 63.7 {\scriptsize\textcolor{violet!80!black}{(+7.7)}} \\
& Qwen2.5-VL-32B & 57.2 & 62.9 {\scriptsize\textcolor{violet!80!black}{(+5.7)}} \\
& Qwen2.5-VL-72B & 58.9 & 65.7 {\scriptsize\textcolor{violet!80!black}{(+6.8)}} \\
\bottomrule
\end{tabular}
\end{table}

\begin{figure*}[!t]
\centering
\includegraphics[width=0.9\textwidth]{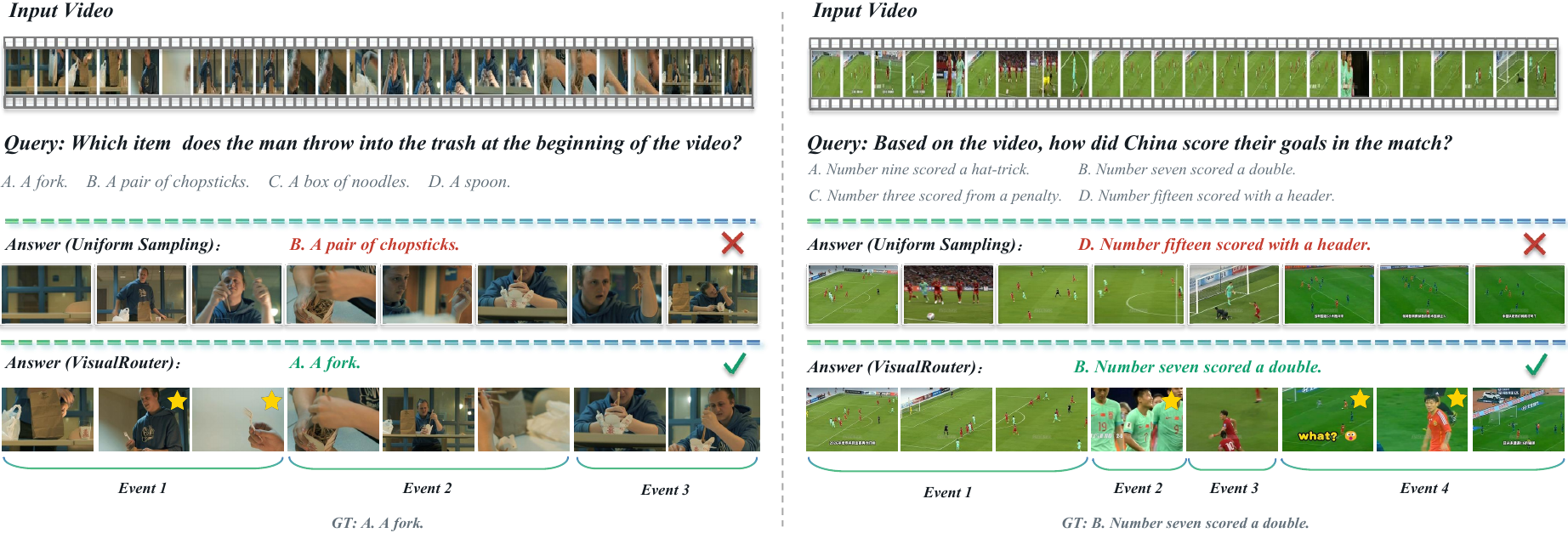}
\caption{Qualitative examples from Video-MME comparing uniform sampling with VisualRouter. In the examples shown, VisualRouter selects frames from relevant event segments and retains visual evidence missed by uniform sampling, resulting in correct predictions.}
\label{fig:visualization_cases}
\end{figure*}

\subsubsection{Effect of LVLM Scale}
We further evaluate VisualRouter on Qwen2.5-VL models ranging from 3B to 72B under a 16-frame budget. As shown in Table~\ref{tab:scale_ablation}, VisualRouter consistently improves uniform sampling across all model scales on both Video-MME and LongVideoBench. Notably, it still yields gains of 3.9 and 6.8 percentage points on the 72B model, respectively. These results demonstrate that effective visual sampling remains complementary to stronger LVLM backbones, rather than benefiting only smaller models.

\subsubsection{Ablation Study of VisualRouter}
Table~\ref{tab:component} evaluates the contribution of each component in VisualRouter. Using global-only branch causes a 4\% drop on LVB, confirming that broad temporal coverage alone is insufficient for questions requiring localized evidence. In contrast, the local-only variant matches the full method on LVB, where all samples are routed to the local branch, but performs slightly worse on Video-MME and MLVU, demonstrating the benefit of query gating on benchmarks with mixed evidence requirements. Random event partitioning and equal frame allocation also degrade performance, with the reducing LVB accuracy by 3.6\%, highlighting the importance of allocating frames according to segment relevance. Finally, replacing K-DPP with Top-$K$ selection leads to small but consistent drops, whereas uniform intra-segment sampling causes substantially larger degradation. These results validate the effectiveness of query gating, frame allocation, and relevance-diversity-aware frame selection. 

\begin{table}[!t]
\caption{Component ablation with Qwen2.5-VL-7B ($K=16$). Each variant
removes or replaces one component while keeping the remaining pipeline
unchanged. Accuracy (\%) is reported.}
\label{tab:component}
\centering
\footnotesize

\resizebox{\columnwidth}{!}{%
\begin{tabular}{@{}lccc@{}}
\toprule
Method & Video-MME & LVB & MLVU \\
\midrule
Uniform                         & 57.1 & 56.0 & 56.3 \\
VisualRouter                             & \textbf{62.3} & \textbf{63.7} & \textbf{67.9} \\
\midrule
\multicolumn{4}{l}{\textbf{Query Gating}} \\
w/o query gating (global only)         & 61.5 & 59.7 & 65.2 \\
w/o query gating (local only)          & 61.8 & 63.7 & 67.4 \\
\midrule
\multicolumn{4}{l}{\textbf{Local Branch}} \\
w/o Event Partitioning (random)          & 62.1 & 62.3 & 67.7 \\
w/o Frame Allocation (equal)  & 61.5 & 60.1 & 66.6 \\
w/o $k$-DPP (Uniform)            & 59.5 & 57.8 & 62.5 \\
w/o $k$-DPP (Top-$K$)            & 62.1 & 63.5 & 67.3 \\
\bottomrule
\end{tabular}%
}
\end{table}

\subsubsection{Ablation of Relevance and Boundary Signals}
Table~\ref{tab:signal_ablation} examines the query-frame relevance and event boundary signals used in VisualRouter. All matching-based variants outperform uniform sampling across the three benchmarks. BLIP2-ITM achieves the best overall performance, tying for the best result on Video-MME and ranking first on LVB and MLVU. We therefore adopt it as the default scorer and use the corresponding visual features for event partitioning. For boundary detection, fusing semantic drift with visual continuity yields the best results on LVB and MLVU while remaining competitive on Video-MME. These results demonstrate that semantic and visual signals capture complementary event changes and enable more reliable event partitioning across benchmarks.

\begin{table}[!t]
\caption{Comparison of relevance scorers and event-boundary signals using
Qwen2.5-VL-7B ($K=16$), with BLIP2-ITM fixed for boundary evaluation.Accuracy (\%) is reported.}
\label{tab:signal_ablation}
\centering
\footnotesize
\begin{tabular}{lccc}
\toprule
Method & Video-MME & LVB & MLVU \\
\midrule
Uniform           & 57.1 & 56.0 & 56.3 \\
\midrule
\multicolumn{4}{l}{\textbf{Query-Frame Relevance}} \\
CLIP~\cite{radford2021clip}             & 62.3 & 56.5 & 65.3 \\
SigLIP~\cite{zhai2023siglip}         & 61.9 & 59.2 & 67.4 \\
BLIP-ITM~\cite{li2022blip}                  & 62.3 & 59.7 & 67.5 \\
BLIP2-ITM (VisualRouter)~\cite{li2023blip2}         & \textbf{62.3} & \textbf{63.7} & \textbf{67.9} \\
\midrule
\multicolumn{4}{l}{\textbf{Boundary Signal for Event Partitioning}} \\
Semantic drift only                         & \textbf{62.5} & 59.6 & 67.6 \\
Visual only                      & 62.4 & 60.2 & 67.6 \\
Semantic + visual fusion (VisualRouter)             & 62.3 & \textbf{63.7} & \textbf{67.9} \\
\bottomrule
\end{tabular}
\end{table}

\subsection{Qualitative Analysis}
Fig.~\ref{fig:visualization_cases} presents two qualitative examples from Video-MME. Uniform sampling selects frames at fixed temporal intervals and may overlook short but query-relevant events, resulting in incorrect answers. In contrast, VisualRouter allocates the limited number of input frames to relevant event segments while preserving representative and diverse frames within each segment. As a result, the LVLM receives more informative visual inputs and produces the correct answers.
\section{Conclusion}
This paper presents VisualRouter, a training-free and plug-and-play framework for query-grounded visual sampling in long video understanding. By routing queries to corresponding visual sampling branches, VisualRouter combines query-relevant evidence with temporal coverage for global queries, while jointly balancing query-frame relevance, evidence coverage, and visual diversity for local queries. This design enables VisualRouter to adapt the sampling strategy to different evidence requirements without retraining. Experiments on Video-MME, LongVideoBench, and MLVU demonstrate consistent improvements across different LVLM architectures, model scales, and numbers of input frames, while achieving competitive performance against representative training-free visual sampling methods under the same evaluation settings.

\section*{Acknowledgments}
This work is supported by the National Key Research and Development Program of China (2025YFF0522500) and the Central Guidance on Local Science and Technology Development Fund of Shanghai City (YDZX20253100002004).

\bibliographystyle{IEEEtran}
\bibliography{TMM}
\end{document}